\documentclass{article}
\usepackage{spconf,amsmath,epsfig}
\usepackage{graphicx}
\usepackage{subfigure}
\usepackage{multirow}
\usepackage{color}
\usepackage{changepage}
\usepackage{times}
\usepackage{float}
\usepackage{amssymb}
\usepackage{multirow}
\usepackage{bbding}
\usepackage{booktabs}
\usepackage{caption}
\usepackage{cite}
\usepackage{stfloats}
\usepackage{url}
\begin{document}
	\title{Combining Visible Light and Infrared Imaging for Efficient Detection of Respiratory Infections such as COVID-19 on Portable Device}
	\name{Zheng Jiang$^{1}$, Menghan Hu$^{2}$, Lei Fan$^{1}$, Yaling Pan$^{3}$, Wei Tang$^{3}$, Guangtao Zhai$^{1}$, and Yong Lu$^{3}$
		\thanks{This work is sponsored by the National Natural Science Foundation of China (No. 61901172, No. 61831015, No. U1908210), the Shanghai Sailing Program (No.19YF1414100),  the STCSM (No.18DZ2270700), the Science and Technology Commission of Shanghai Municipality (No. 19511120100), and the foundation of Key Laboratory of Articial Intelligence, Ministry of Education (No. AI2019002).}
		\thanks{Z. Jiang and M. Hu  contributed equally to this work.}
		\thanks{Corresponding authors: Guangtao Zhai (zhaiguangtao@sjtu.edu.cn), Yong Lu (18917762053@163.com)}
	}
\address{$^1$Institute of Image Communication and Information Processing, Shanghai Jiao Tong University, China\\
	$^2$Shanghai Key Labora. of Multidim. Infor. Proce., East China Normal University, China\\
		$^{3}$ Rui Jin Hospital/Lu Wan Branch, School of Medicine, Shanghai Jiao Tong University\\
}

	\maketitle
	
	\begin{abstract}
		Coronavirus  Disease 2019 (COVID-19) caused by severe acute respiratory syndrome coronaviruses 2 (SARS-CoV-2) has become a serious global epidemic in the past few months and caused huge loss to human society worldwide. 
		For such a large-scale epidemic, early detection and isolation of potential virus carriers is essential to curb the spread of the epidemic. Recent studies have shown that one important feature of COVID-19 is the abnormal respiratory status caused by viral infections. During the epidemic, many people tend to wear masks to reduce the risk of getting sick. Therefore, in this paper, we propose a portable non-contact method to screen the health condition of people wearing masks through analysis of the respiratory characteristics. The device mainly consists of a FLIR one thermal camera and an Android phone. This may help identify those potential patients of COVID-19 under practical scenarios such as pre-inspection in schools and hospitals. 
		In this work, we perform the health screening through the combination of the RGB and thermal videos obtained from the dual-mode camera and deep learning architecture.
		We first accomplish a respiratory data capture technique for people wearing masks by using face recognition. Then, a bidirectional GRU neural network with attention mechanism is applied to the respiratory data to obtain the health screening result. The results of validation experiments show that our model can identify the health status on respiratory with the accuracy of 83.7\% on the real-world dataset. The abnormal respiratory data and part of normal respiratory data are collected from Ruijin Hospital Affiliated to The Shanghai Jiao Tong University Medical School. Other normal respiratory data are obtained from healthy people around our researchers. This work demonstrates that the proposed portable and intelligent health screening device can be used as a pre-scan method for respiratory infections, which may help fight the current COVID-19 epidemic. The demo videos of the proposed system are available at: https://doi.org/10.6084/m9.figshare.12028032. 
	\end{abstract}
	
	
	\section{Introduction}
	\label{sec:introduction}
	During the outbreak of COVID-19 epidemic, early control is essential. Among all the control measures, efficient and safe identification of potential patients is the most important part. Existing researches show that human physiological state can be perceived through breathing \cite{cretikos2008respiratory}, which means respiratory signals are vital signs that can reflect human health condition to a certain extent\cite{droitcour2009non}. Many clinical literature suggests that abnormal respiratory symptoms may be important factors for diagnosis of some specific diseases\cite{boulding2016dysfunctional}.
	Recent studies have found that COVID-19 patients will have obvious respiratory symptoms such as shortness of breath fever, tiredness, and dry cough\cite{xu2020pathological,sohrabi2020world}. Among those symptoms, atypical or irregular breathing is considered as one of the early signs. For many people, early mild respiratory symptoms are difficult to be recognized. Therefore, through the measurement of respiration condition, potential COVID-19 patients can be screened to some extent. This may play an auxiliary diagnostic role, thus helping to find potential patients as early as possible.
	
	Traditional respiration measurement requires attachments of sensors to the patient's body\cite{al2011respiration}. The monitor of respiration is measured through the movement of the chest or abdomen. Contact measurement equipment is bulky, expensive, and time-consuming. The most important thing is that the contact during measurement may increase the risk of spreading infectious diseases such as COVID-19. Therefore, the non-contact measurement is more suitable for the current situation. In recent years, many non-contact respiration measurement methods have been developed based on imaging sensors, doppler radar\cite{kranjec2013novel}, depth camera\cite{wang2020abnormal} and thermal camera\cite{hu2017synergetic}. Considering factors such as safety, stability and price, the measurement technology of thermal imaging is the most suitable for extensive promotion. So far, thermal imaging has been used as a monitoring technology in a wide range of medical fields such as estimations of heart rate\cite{hu2018combination} and breathing rate\cite{pereira2015remote,lewis2011novel,chen2019rgb}. 
	Another important thing is that many existing respiration measurement devices are large and immovable. Given the worldwide epdemic, the partable and intelligent screening equipment is required to meet the needs of large-scale screening and other application scenarios in a real-time manner. For thermal imaging based respiration measurement, nostril regions and mouth regions are the only focused regions since only these two parts have periodic heat exchange between the body and the outside environment. However, until now, researchers seldom considered measuring thermal respiration data for people wearing masks. During the epidemic of infectious diseases, masks may effectively suppress the spread of the virus according to recent studies\cite{feng2020rational,leung2020mass}. Therefore, developing the respiration measurement method for people wearing masks becomes quite practical. 
	In this study, we develop a portable and intelligent health screening device that uses thermal imaging to extract respiration data from masked people which is then used to do the health screening classification via deep learning architecture.
	
	In classification tasks, deep learning has achieved the state-of-the-art performance in most research areas. Compared with traditional classifiers, classifiers based on deep learning can automatically identify the corresponding features and their correlations rather than extracting features manually. For breathing tasks, algorithms based deep learning can also better extract the corresponding features such as breathing rate and breath-to-exhale ratio, and make more accurate predictions\cite{chauhan2018performance,liu2018deep,zhang2017respiration,khan2017deep}. Recently, Many researchers made use of deep learning to analyze the respiratory process. Cho et al. used a convolutional neural network (CNN) to analyze human breathing parameters to determine the degree of nervousness through thermal imaging\cite{cho2017deepbreath}. Romero et al. applied a language model to detect acoustic events in sleep-disordered breathing through related sounds\cite{8683099}. Wang et al. utilized deep learning and depth camera to classify abnormal respiratory patterns in real time and achieved excellent results\cite{wang2020abnormal}. The disadvantage of this research may be that the equipment is not portable.

	In this paper, we propose a remote, potable and intelligent health screening system based on respiratory data for pre-screening and auxiliary diagnosis of respiratory diseases like COVID-19. In order to be more practical in a situation where people often choose to wear masks, the breathing data capture method for people wearing masks is introduced. After extracting breathing data from the video obtained from the thermal camera, a deep learning neural network is performed to work on the  classification between healthy and abnormal respiration conditions. To verify the robustness of our algorithm and the effectiveness of the proposed equipment, we analyze the influence of mask type, measurement distance and measurement angle on breathing data.

	The main contributions of this paper are threefold. First, we combine the face recognition technology with dual-mode imaging to accomplish a respiratory data extraction method for people wearing masks, which is quite essential for current situation. Based on our dual-camera algorithm, the respiration data is successfully obtained from masked facial thermal videos. Subsequently, we propose a classification method to judge abnormal respiratory state with deep learning framework. Finally, based on the two contributions mentioned above, we have implemented a non-contact and efficient health screening system for respiratory infections using the actual measured data from hospital, which may contribute to finding the possible cases of COVID-19 and keeping the control of the secondary spread of SARSCoV-2.

	\section{Method}
	A brief introduction to the proposed respiration condition screening method is shown below. We first use the portable and intelligent screening device to get the thermal and the corresponding RGB videos. During the data collection, we also perform a simple real-time screening result. After getting the thermal videos, the first step is to extract respiration data from faces in thermal videos. During the extraction process, we use the face detection method to capture people's masked areas. Then a region of interest (ROI) selection algorithm is proposed to get the region from the mask that stands for the characteristic of breath most. Finally, we use a bidirectional GRU neural network with attention mechanism (BiGRU-AT) model to work on the classification task with the input respiration data. 
	
	\subsection{Overview of Portable and Intelligent Health Screening System for Respiratory Infections}
\begin{figure}[ht]
		\centering
		\includegraphics[width=0.4\textwidth]{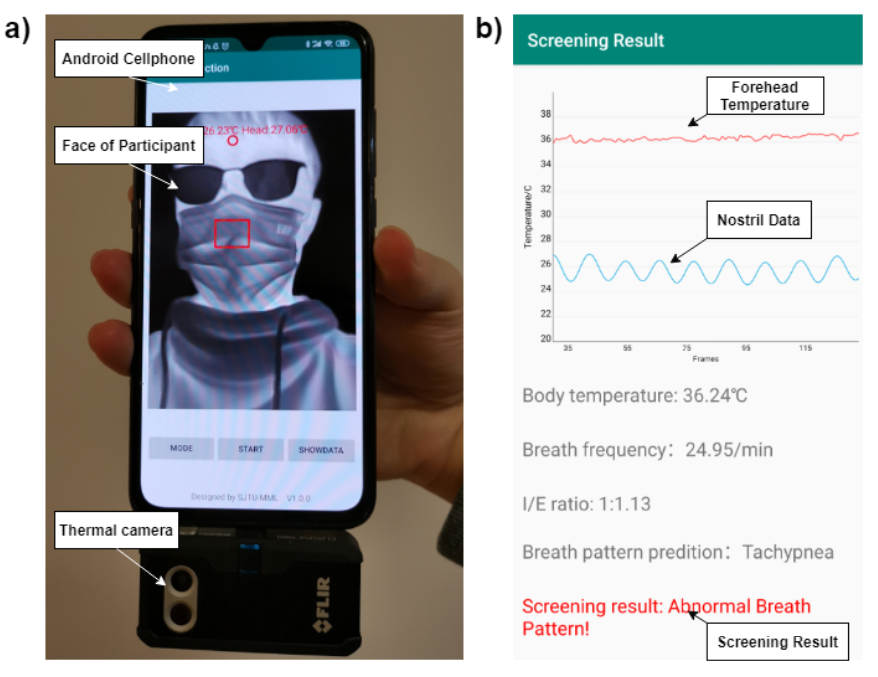}
		\caption{Overview of portable and intelligent health screening system for respiratory infections: a) device appearance; b) analysis result of the application. (Notice that the system can simultaneously collect body temperature signals. In the current work, this body temperature signal is not considered in the model and is only used as a reference for the users. )}\label{device}
	\end{figure}

	Our data collection is achieved by the system shown in Fig. \ref{device}. The whole screening system includes a FLIR one thermal camera, an Android smartphone and the corresponding application we have written, which is used for data acquisition and simple instant analysis. Our screening equipment, whose main advantage is portable, can be easily applied to measure abnormal breathing in many occasions of instant detection.

	As shown in Fig. \ref{device}, the FLIR one thermal camera consists of two canmeras, an RGB camera and a thermal camera. We collect the face videos from both cameras and use face recogition method to get the nostril area and forehead area. The temperatures of the two regions are calculated in time series and is shown in the screening result page in Fig. \ref{device}(b). The red line stands for the body temprature and the blue line stands for breathing data. From the breathing data, we can predict the respiratory pattern of the testcase. Then, a simple real-time screening result is given directly in the application according to the extracted features shown in Fig. \ref{device}. 
	We use the the raw face videos collected from both RGB camera and thermal camera as the data for further study to ensure accuracy and higher performance.
	
	\subsection{Detection of Masked Region from Dual Mode Image}
	
	When continuous breathing activities performs, there is a fact that periodic temperature fluctuations occur around the nostril due to the inspiration and expiration cycles. Therefore, respiration data can be obtained by analyzing the temperature data around the nostril based on thermal image sequence. However, when people wear masks, many facial features are blocked because of this. Merely recognizing the face through thermal image will lose a lot of geometric and textural facial details, resulting in recognition errors of the face and mask parts. In order to solve this problem, we adopt the method based on two parallel located RGB and infrared cameras for face and mask region recognition. The masked region of face is first captured in the RGB camera, then such region is mapped to the thermal image with a related mapping function.
	
	The algorithm for masked face detection is based on pyramidbox model created by Tang et al.\cite{tang2018pyramidbox}. The main idea is to apply tricks like Gaussian pyramidbox in deep learning to get the context correlations as further characteristics. The face image is first used to extract features of different scales using Gaussian pyramid algorithm. For those high-level contextual features, a feature pyramid network is proposed to further excavate high-level contextual features. Then, the output together with those low-level features are combined in low-level feature pyramid layers. Finally, the result is obtained after another two layers of deep neural network. For faces that a lot of features are lost due to the cover of mask, such a context-sensitive structure can obtain more feature correlations and thus improve the accuracy of face detection.
	In our experiment, we use the open source model from paddlehub to detect the face area on RGB videos. 
	
	The next step is to extract the masked area and map the area from RGB video to thermal video. Since the position of the mask on the human face is fixed, after obtaining the position coordinates of the human face, we obtain the mask area of the face by scaling down in equal proportions. For a detected face with width $w$, and height $h$, the loaction of left-up corner is defined as $(0, 0)$, the loaction of right-bottom corner is then $(w, h)$. The corresponding coordinate of the two corners of mask region is decalred as $(w/4, h/2)$ and $(3w/4, 4h/5)$. Considering that the background to the boundary of the mask will produce a large contrast with the movement, which is easy to cause errors, we choose the center area of the mask through this division.
	Then the selected area is mapped from the RGB image to thermal image to obtain the masked region in thermal videos.
	
	\begin{figure*}[ht]
	\centering
	\includegraphics[width=0.9\textwidth]{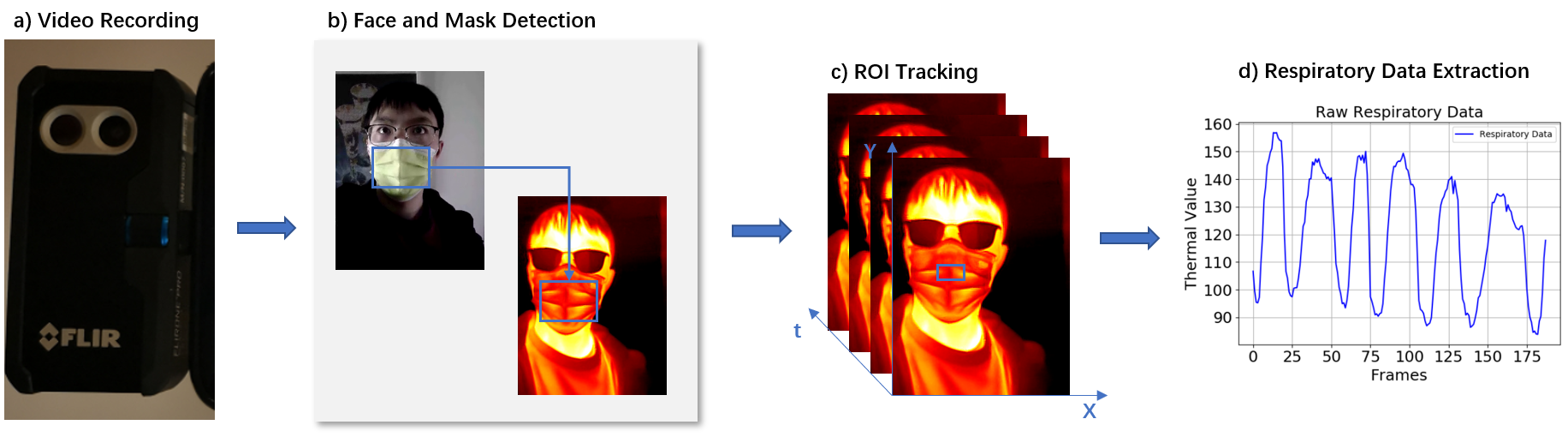}
	\caption{The pipeline of respiration data extraction: a) record the RGB video and thermal video through a FLIR one thermal camera; b) use face detection method to detect face and mask region in the RGB frames and then map the region to the thermal frames; c) capture the ROIs in the thermal frames of mask region by tracking method; d) extract the respiration data from the ROIs.}\label{F01}
\end{figure*}
	\subsection{Extract Respiration Data from ROI}
	After getting the masked region in thermal videos, we need to get the region of interest (ROI) that represents breathing features. Recent studies often characterize breathing data through temperature changes around the nostril\cite{hu2018combination,cho2017robust}.
	However when people wear masks, there exists another problem that the nostrils are also blocked by the masks, and when people wearing different masks, the ROI may be different. Therefore, we perform a ROI tracking method based on maximizing the variance of thermal image sequence to extract a certain area on the masked region of the thermal video which stands for the breath signals most.
	
	Due to the lack of texture features in masked regions compared to human faces, we judge the ROI by the temperature change of thermal image sequence. The main idea is to traverse the masked region in the thermal images and find a small block with the largest temperature change as the selected ROI. The position of a certain block is fixed in the masked region among all the frames since the nostril area is fixed on the face region. We do not need to consider the movement of the block since our face recognition algorithm can detect the mask position in each frame's thermal image. For a certain block with height $m$, and width $n$, we define the average pixel intensity at frame $t$ as:
	
	\begin{equation}
	\bar{s}(t)=\frac{1}{m n} \sum_{i=0}^{m-1} \sum_{j=0}^{n-1} s(i, j,t)
	\end{equation}
	
	For thermal images, $\bar{s}(t)$ represents the temperature value at frame $t$. For every block we obtained, we calculate their $\bar{s}(t)$ on time line. Then, for each block $n$, the total variance of the list of average pixel intensity with $T$ frames $\sigma_{s}^{2}(n)$ is calculated as shown in Eq. \ref{variance}, where $\mu$ stands for the mean value of $\bar{s}(t)$.
	\begin{equation}
	\sigma_{s}^{2}(n)=\frac{\sum(\bar{s}(t)-\mu)^{2}}{T}(0<t<T)
	\label{variance}
	\end{equation}
	
	Since respiration is a periodic data spread out from the nostril area, we can consider that the block with the largest variance is the position where the heat changes most in both frequency and value within the mask, which stands for the breath data most in the masked region. We adjust the corresponding block size according to the size of the masked region. For a masked region with $N$ blocks, the final ROI is selected by:

	\begin{equation}
	ROI=\underset{1 \leq n<N}{\arg \max }{\sigma_{s}^{2}(n)}
	\end{equation}
	
	For each thermal video, we traverse all possible blocks in the mask regions of each frame and find the ROIs for each frame by the method above. The respiration data is then defined as $\bar{s}(t)(0<t<T)$, which is the pixel intensities of ROIs in all the frames.
	
		\begin{figure*}[ht]
		\centering
		\includegraphics[width=0.9\textwidth]{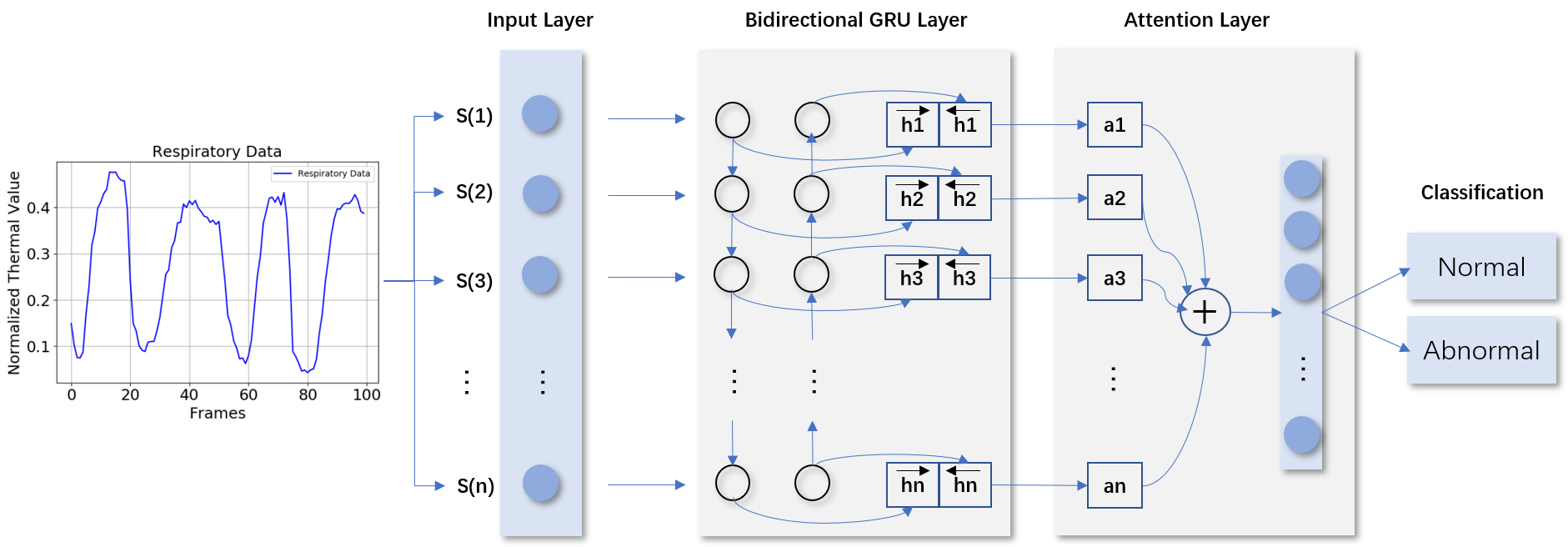}
		\caption{The structure of the BiGRU-AT network: the network consists of four layers: the input layer, the bidirectional layer, the attention layer and a final dense layer. The output is a 2 dimension tensor which indicates normal or abnormal respiration condition.}\label{F02}
	\end{figure*}

	\subsection{BiGRU-AT Neural Network}
	
	We apply a BiGRU-AT neural network to do the classification task on judging whether the respiration condition is healthy or not as shown in Fig. \ref{F02}. The input of the network is the respiration data obtained by our extraction method. Since the respiratory data is time series, it can be regarded as a time series classification problem. Therefore, we choose the Gate Recurrent Unit (GRU) network with bidirection and attention layer to work on the sequence prediction task. 
	
	Among all the deep learning structures, recurrent neural network (RNN) is a type of neural network which is specially used to process time series data samples\cite{elman1990finding}. For a time step $t$, the RNN model can be represented by:
	
	\begin{equation}h^{(t)}=\phi\left(U x^{(t)}+W h^{(t-1)}+b\right)\end{equation}
	\begin{equation}o^{(t)}=V h^{(t)}+c\end{equation}
	\begin{equation}\widehat{y}^{(t)}=\sigma\left(o^{(t)}\right)\end{equation}
	where $x^{(t)}, h^{(t)}$ and  $o^{(t)}$ stands for the current input state, hidden state and output at time step $t$ respectively. $V, W, U$ are parameters obtained by training procedure. $b$ is the bias and $\sigma$ and $\phi$ are activation functions. The final prediction is $\widehat{y}^{(t)}$.
	
	Long-short term memory network is developed on the basis of RNN\cite{hochreiter1997long}. Compared to RNN, which can only memorize and analyze short-term information, it can process relatively long-term information, and is suitable for problems with short-term delays or long time intervals. 
	Based on LSTM, many related structures are proposed in recent years\cite{greff2016lstm}. GRU is a simplified LSTM which merges three doors of LSTM (forget, input and output) into two doors (update and reset)\cite{cho2014learning}.
	For tasks with a few data, GRU may achieve a better result than LSTM since it includes less parameters. In our task, since the input of the neural network is only the respiration data in time sequence, the GRU network may perform a better result than LSTM network. The structure of GRU can be expressed by the following
	equations:
	
	\begin{equation}r_{t}=\sigma\left(W_{r} \cdot\left[h_{t-1}, x_{t}\right]+b_r\right)\end{equation}
	\begin{equation}
	z_{t}=\sigma\left(W_{z} \cdot\left[h_{t-1}, x_{t}\right]+b_{z}\right)
	\end{equation}
	\begin{equation}\tilde{h}_{t}=\tanh \left(W_{\bar{h}} \cdot\left[r_{t} * h_{t-1}, x_{t}\right]+b_h\right)\end{equation}
	\begin{equation}h_{t}=\left(1-z_{t}\right) * h_{t-1}+z_{t} * \tilde{h}_{t}\end{equation}
	where $r_t$ is the reset gate that controls the amount of information being passed to the new state from the previous states. $z_t$ stands for the update gate which determines the amount of information being forgotten and added. $W_{r},W_{z}$ and $W_{h}$ are trained parameters that vary in the training procedure. $\tilde{h}_{t}$ is the candidate hidden layer which can be regarded as a summary of the above information $h_{t-1}$ and the input information $x_t$  at time step $t$.
	$h_t$ is the output layer at time step $t$ which will be sent to the next unit. 
	
	The bidirectional RNN has been widely used in natural language processing \cite{bahdanau2014neural}. The advantage of such network structure is that it can strengthen the correlation between context of the sequence. As the respiratory data is a periodic sequence, we use bidirectional GRU to obtain more information from the periodic sequence. The difference between bidirectional GRU and normal GRU is that backfoward sequence of data is spliced to the original forward sequence of data. In this way, the hidden layer of the original $h(t)$ is changed to:
	\begin{equation}h_{t}=[\overrightarrow{h_{t}}, \overleftarrow{h_{t}}]\end{equation}
	where $\overrightarrow{h_{t}}$ is the original hidden layer and $\overleftarrow{h_{t}}$ is the backfoward sequence of  $\overrightarrow{h_{t}}$. 
	
	During the analysis of respiratory data, the entire waveform in time sequence should be taken into consideration. 
	For some speciﬁc breathing pattern such as sphyxia, several paticular features such as sudden acceleration may occur only at a certain point in the entire process.
	However, if we only use the BiGRU network, these features may be weakened as the time sequence data is input step by step which may cause a larger error in prediction. Therefore, we add an attention layer to the network, which can ensure that certain key point features in the breathing process can be maximized.
	
	Attention mechanism is a choice to focus only on those important points among the total data\cite{vaswani2017attention}. It is often combined with neural networks like RNN. Before the RNN model summarizes the hidden states for the output, an attention layer can make an estimation of all outputs and find the most important ones. This mechanism has been widely used in many research areas. The structure of attention layer is:
	\begin{equation}u_{t}=\tanh \left(W_{u} h_{t}+b_{w}\right)\end{equation}
	\begin{equation}a_{t}=\frac{\exp \left(u_{t}^{\top} u_{w}\right)}{\sum_{t} \exp \left(u_{t}^{\top} u_{w}\right)}\end{equation}
	\begin{equation}s=\sum_{t} \alpha_{t} h_{t}\end{equation}
	where $h_t$ represents the BiGRU layer output at time step $t$, which is bidirectional. $W_u$ and $b_w$ are also parameters that vary in the training process. $a_t$ performs a softmax fucntion on $u_t$ to get the weight of each step $t$. Finally, the output of the attention layer $s$ is a combination of all the steps from BiGRU with different weights. By applying another softmax function to the output $s$, we get the final prediction of the classification task. The structure of the whole network is shown in Fig. \ref{F02}.
	\section{Experiments}
	
	\subsection{Dataset Explanation and Experimental Settings}
	Our goal is to distinguish whether there is an epidemic infectious disease such as COVID-19 according to the abnormal breathing in the respiratory system. 
	Our dataset is not collected from patients with COVID-19.
	These data were obtained from the inpatients of the respiratory disease department and cardiology department in Ruijin Hospital. Most of the patients we collected data from only caught with basic or chronic respiratory disease. They did not have fever which is the typical respiratory symptoms of infectious diseases. Therefore, the body temperature is not taken into consideration in our current screening system. 
	
	In Ruijin Hospital wards, we use a FLIR one thermal camera connected to an Android phone to work on the data collection. We collected data from 73 people. For each person, we collect two 20-second infrared and RGB camera data with a sampling frequency of 10 Hz. Through data cutting and oversampling, we finally obtained 1,925 healthy breathing data and 2,292 abnormal breathing data, a total of 4,217 data. Each piece of data consists of 100 frames of infrared and RGB videos in 10 seconds. 
	In the BiGRU-AT network, the hidden cells in BiGRU layer and attention layers are 32 and 8 respectively. The breathing data is normalized before input into the neural network and we use cross-entropy as the loss function. During the training process, we separate the dataset into two parts. The training set includes 1,427 healthy breathing data and 1,780 abnormal breathing data. And the test set contains 498 healthy breathing data and 512 abnormal breathing data. Once this paper is accepted, we will release the dataset used in the current work for non-commercial users. 
	
	\begin{figure}[ht]
		\centering
		\includegraphics[width=0.5\textwidth]{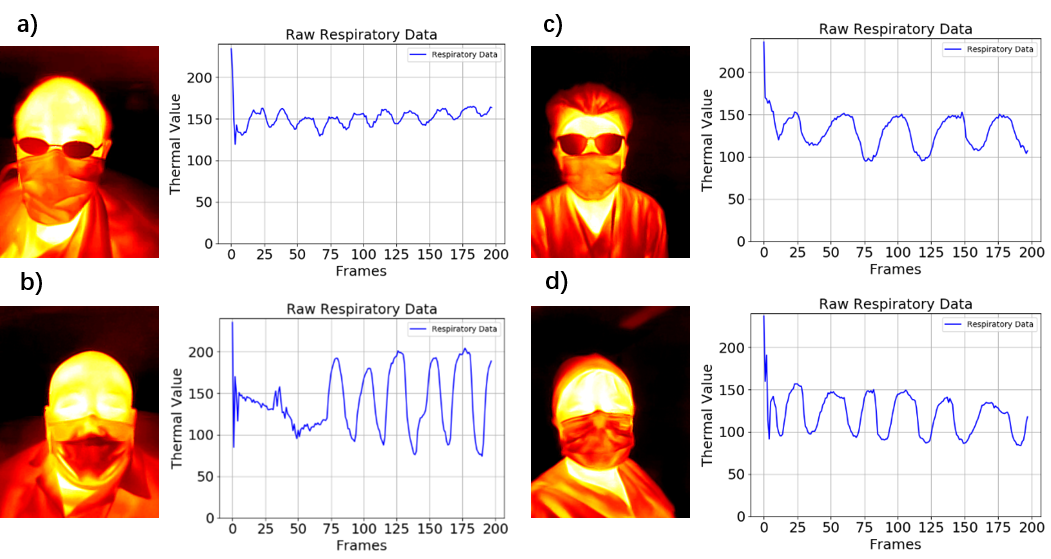}
		\caption{A Comparison of normal and abnormal respiratory data: a), b) are abnormal data collected from patients in the general ward of the respiratory department in Ruijin Hospital; c), d) are normal data collected from healthy volunteers.}\label{sick}
	\end{figure}
	Among the whole dataset, we choose 4 typical respiratory data examples as shown in Fig. \ref{sick}.  Fig. \ref{sick}(a) and Fig. 4(b) stand for the abnormal respiratory patterns extracted from patients. Fig. \ref{sick}(c) and Fig. \ref{sick}(d) represent the normal respiratory pattern called Eupnea from healthy participants. By comparison, we can find that the respiratory of normal people is in strong periodic and evenly distributed while abnormal respiratory data tend to be more irregular. Generally speaking, most abnormal breathing data from respiratory infections have faster frequency and irregular amplitude.
	
	\subsection{Experimental Result}
	The experimental results are shown in Table. \ref{tab1}. We consider four evaluation metrics viz. Accuracy, Precision, Recall and F1. To measure the performance of our model, we compare the result of our model with three other models which are GRU-AT, BiLSTM-AT and LSTM respectively. 
	The result shows that our method performs better than any other networks in all evaluation metrics despite the precision value of GRU-AT. By comparison, the experimental result demonstrates that attention mechanism is well-performed in keeping important node features in the time series of breathing data since the networks with attention layer all perform a better result than LSTM. Another point is that GRU based networks achieve better result than LSTM based networks. This may beacuse our data set is relatively small which can't fill the demand of the LSTM based networks. GRU based networks require less data than LSTM and perform better result in our respiration condition classification task. 
	\begin{table}[!hpb]
		\centering
		\caption{Exprimental results on the test set}
		\setlength{\tabcolsep}{2mm}{
			\begin{tabular}{ccccc} 
				\hline
				Model&Accuracy &Precision&Recall &F1\%\\
				\hline 
				BiGRU-AT & 83.69\% & 90.23\%&79.65\% & 84.61\%\\
				GRU-AT & 79.31\% & 90.62\%&74.24\% & 81.62\%\\
				BiLSTM-AT & 74.46\% & 87.50\% & 69.78\% & 77.64\%\\
				LSTM & 71.98\% & 72.07\% & 71.98\%  & 71.97\%\\
				\hline
		\end{tabular}}
		\label{tab1}
	\end{table}
	
\begin{figure}[!htp]
		\begin{minipage}{0.48\linewidth}
			\centerline{\includegraphics[width=1\textwidth]{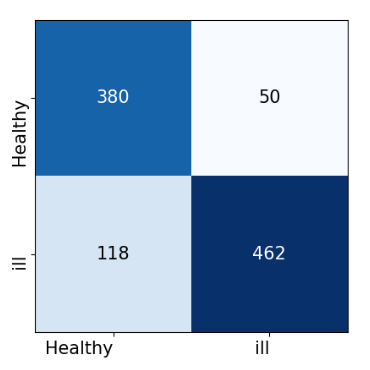}}
			\centerline{(a) BiGRU-AT}
		\end{minipage}
		\hfill
		\begin{minipage}{0.48\linewidth}
			\centerline{\includegraphics[width=1\textwidth]{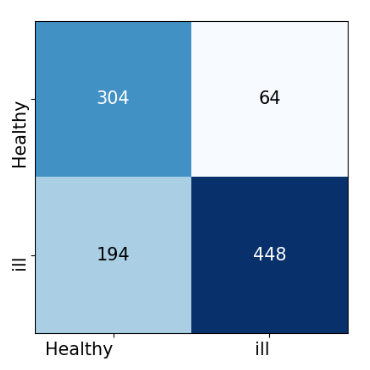}}
			\centerline{(b) BiLSTM-AT}
		\end{minipage}
		\vfill
		\begin{minipage}{0.48\linewidth}
			\centerline{\includegraphics[width=1\textwidth]{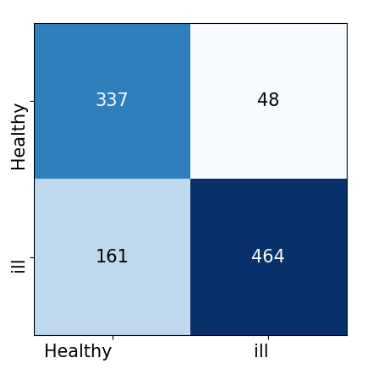}}
			\centerline{(c) GRU-AT}
		\end{minipage}
		\hfill
		\begin{minipage}{0.48\linewidth}
			\centerline{\includegraphics[width=1\textwidth]{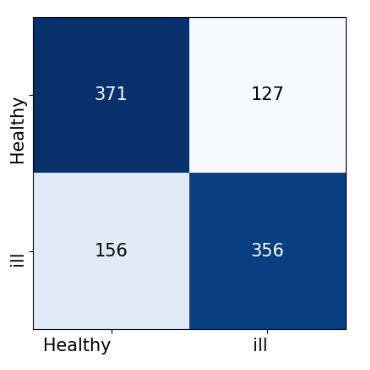}}
			\centerline{(d) LSTM}
		\end{minipage}
		\centering
		\caption{Confusion matrices of the four models. Each row is the number
			of real labels and each column is the number of predicted
			labels. The left one is the result of BiGRU-AT, the right one is the result of LSTM.}
		\label{fig:4}
	\end{figure}

	To figure out the detailed information about the classification of respiratory state, we plotted the confusion matrixs of the four models as demonstrated in Fig. \ref{fig:4}. As can be seen from the results, the performance improvement of BiGRU-AT compared to LSTM is mainly in the accuracy rate of the negative class. This is because many scatter-like abnormalities in the time series of abnormal breathing are better recognized by the attention mechanism. Besides, the misclassification rate of the four networks are relatively high to some extent which may because many positive samples do not have typical respiratory infections characteristics since part of the patients caught other lung-related diseases.
	
	\subsection{Analysis}
	In the analysis section, we give 3 comparasions from different aspects to prove the robustness of our algorithm and device.
	\subsubsection{Influence of Mask Types on Respiratory Data}
	In order to measure the robustness of our breathing data acquisition algorithm and the effectiveness of the proposed portable device, we analyze the breathing data of the same person wearing different masks. We design 3 mask wearing scenarios that cover most situations: wearing one surgical mask (blue line); wearing one KN95 (N95) mask (red line) and wearing two surgical masks (green line). 
	\begin{figure}[ht]
		\centering
		\includegraphics[width=0.45\textwidth]{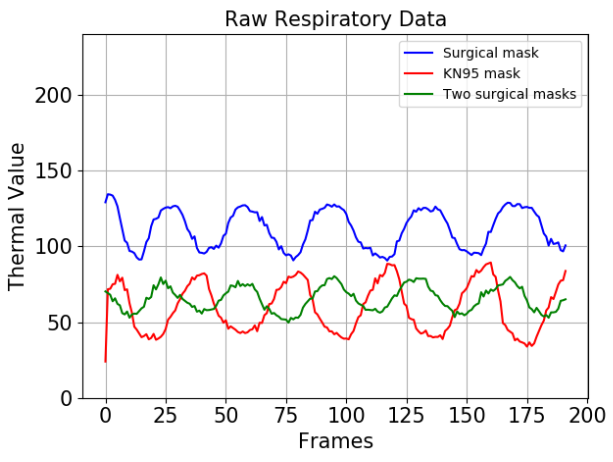}
		\caption{The raw respiratory data obtained through the breathing data extraction algorithm with different types of masks.}\label{mask}
	\end{figure}
	The results are shown in Fig. \ref{mask}. It can be seen from the experimental results that no matter what kind of mask is worn, or even two masks, the respiratory data can be well recognized. This proves the stability of our algorithm and device. However, since different masks have different thermal insulation capabilities, the average breathing temperature may vary as the mask changes. To minimize this error, respiratory data are normalized before input into the neural network. 
	\begin{figure}[ht]
		\centering
		\includegraphics[width=0.45\textwidth]{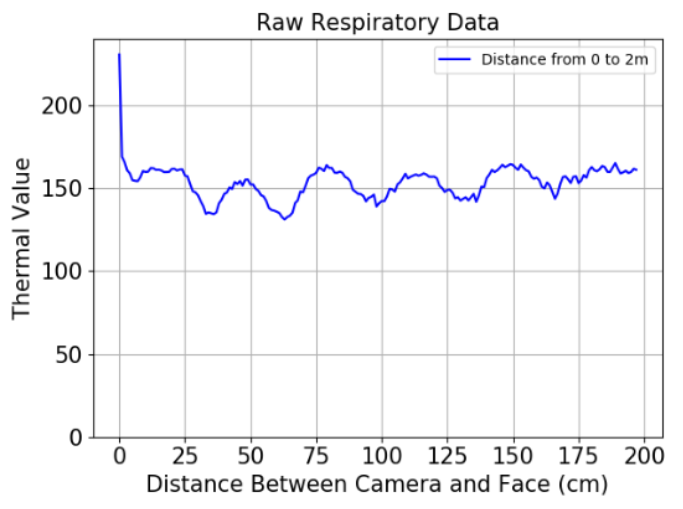}
		\caption{The raw respiratory data which is obtained under the distance between camera and device from 0 to 200 cm.}\label{distance}
	\end{figure}
	\subsubsection{Limitation of Distance to the Camera during Measurement} In order to verify the robustness of our algorithm and device in different scenarios, we design experiments to collect respiratory data at different distances. Considering the limitations of handheld devices, we test the collection of facial respiration data from a distance of 0 to 2 meters. The result is demonstrated in Fig. \ref{distance}. The signal tends to be periodic from the position of 10 centimeters, and it does not start to lose regularity until about 1.8 meters. At a distance of about 10 centimeters, the complete face begins to appear in the camera video. When the distance comes to 1.8 meters, our face detection algorithm begins to fail gradually due to the distance and pixel limitation. This experiment verifies that our algorithm and device can guarantee relatively accurate measurement results in the distance range of 0.1 meters to 1.8 meters.
	
	\subsubsection{Limitation of Angle to the Camera during Measurement} Considering that breath detection will be applied in different scenarios, we design this experiment to show the actual situation under different shooting angles. We define the camera directly towards the face to be 0 degree, and design an experiment in which the shooting angle gradually changed from 45 degrees to 0 degree. We consider the transformation of two angles: horizontal and vertical, which respectively represent left and right turning and nodding.
	\begin{figure}[ht]
		\centering
		\includegraphics[width=0.45\textwidth]{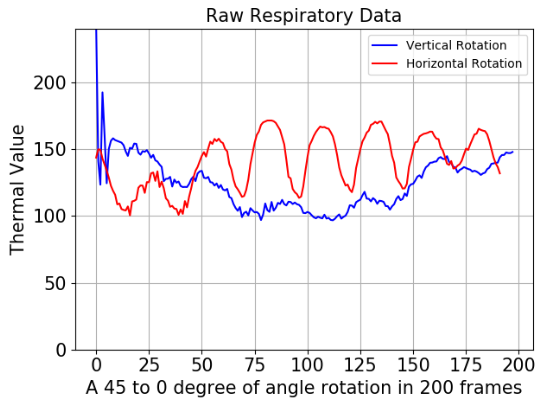}
		\caption{The raw respiratory data obtained while the rotation from 45 degree of angle to 0 degree of angle. The blue line stands for the vertical rotation and the red line stands for the horizontal rotation.}\label{rotation}
	\end{figure}
	The results in the two cases are quite different as shown in Fig. \ref{rotation}. Our algorithm and device maintain good results in horizontal rotation, but it is difficult to obtain precise respiratory data in vertical rotation. This means participants can trun left or turn right during the measurement but can't nod or head up since this may impact the measurement result.
	\section{Conclusion}
	In this paper, we propose a abnormal breathing detection method based on a portable dual-mode camera which can reocrd both RGB and thermal videos. 
	In our detection method, we first accomplished an accurate and robust respiratory data detecion algorithm which can precisely extract breathing data from people wearing masks. Then, a BiGRU-AT network is applied to work on the screening of respiratory infections. In validation experiments, the obtained BiGRU-AT network achieves a realtively good result with the accuracy of 83.7\% on the real-world dataset. It is foreseeable that in patients with COVID-19 who have more clinical respiratory symptoms, this classification method may yield better results. 
	During the current outbreak of COVID-19, our research can work as prescan method for abnomral breathing in many scenarios such as community, campus and hospital which may contribute to distuigishing the possible cases , and then keep the control of the virus spread.

	In future research, on the basis of ensuring portability, we plan to use a more stable algorithm to minimize the effect of different masks on measurement of breathing condition. Besides, temperature may be taken into consideration to achieve a higher detection accuracy on respiratory infections.


	{
		\bibliographystyle{IEEEbib}
		\small
		\bibliography{IEEEabrv,refs}
	}

\end{document}